\documentclass{article}
\usepackage[english]{babel}
\usepackage[letterpaper,top=2cm,bottom=2cm,left=3cm,right=3cm,marginparwidth=1.75cm]{geometry}

\usepackage{authblk}
\usepackage{amsmath}
\usepackage{graphicx}
\usepackage{multirow}
\usepackage{subcaption}
\usepackage[colorlinks=true, allcolors=blue]{hyperref}
\usepackage{comment}
\usepackage{array}

\title{Reducing Activation Recomputation \\ in Large Transformer Models}
\author{Vijay Korthikanti, Jared Casper, Sangkug Lym, Lawrence McAfee, Michael Andersch, Mohammad Shoeybi, and Bryan Catanzaro}
\affil{NVIDIA}

\date{}

\begin{document}
\maketitle

\begin{abstract}
Training large transformer models is one of the most important computational challenges of modern AI. In this paper, we show how to significantly accelerate training of large transformer models by reducing activation recomputation. Activation recomputation is commonly used to work around memory capacity constraints. Rather than storing activations for backpropagation, they are traditionally recomputed, which saves memory but adds redundant compute. In this work, we show most of this redundant compute is unnecessary because we can reduce memory consumption sufficiently without it. We present two novel yet very simple techniques: sequence parallelism and selective activation recomputation. In conjunction with tensor parallelism, these techniques almost eliminate the need to recompute activations. We evaluate our approach on language models up to one trillion parameters in scale and show that our method reduces activation memory by $5\times$, while reducing execution time overhead from activation recomputation by over 90\%. For example, when training a 530B parameter GPT-3 style model \cite{mt-nlg} on 2240 NVIDIA A100 GPUs, we achieve a Model Flops Utilization of 54.2\%, which is 29\% faster than the 42.1\% we achieve using recomputation. Our implementation will be available in both Megatron-LM\footnote{\url{https://github.com/NVIDIA/Megatron-LM} Megatron-LM is an ongoing research project pushing the limits of scale for transformer models.} and NeMo-Megatron\footnote{\url{https://developer.nvidia.com/nvidia-nemo}. NeMo-Megatron is NVIDIA's officially supported product for training large language models.}.

\end{abstract}

\section{Introduction}
\label{sec:intro}

As transformer models scale towards trillions of parameters, model parallelism is required to distribute model parameters, activations, and optimizer state across devices for them to fit into device memory and be trainable in a realistic amount of time. Although model parallelism linearly reduces the number of parameters per device, e.g., number of parameters per device is halved when model parallel size is doubled, there are limits to scale model parallelism. Tensor-level model parallelism increases communication requirements and introduces smaller and less performant matrix multiplications, making it inefficient to split a model across a large number of devices. As a result, tensor-level model parallelism is typically limited to a relatively small group of GPUs that are connected with high speed bandwidth, such as GPUs connected with NVLink inside a DGX server. Pipeline parallelism requires storing the activations of several microbatches to reduce the pipeline bubble \cite{megatron-pipeline}. As a result, pipeline parallelism can only help with the memory needed to store model parameters and optimizer state and cannot reduce the memory needed for activations while maintaining high device utilization. Thus the storage of activations quickly becomes a critical problem to scaling large transformer models.

\begin{figure}
\begin{center}
  \includegraphics[scale=0.3]{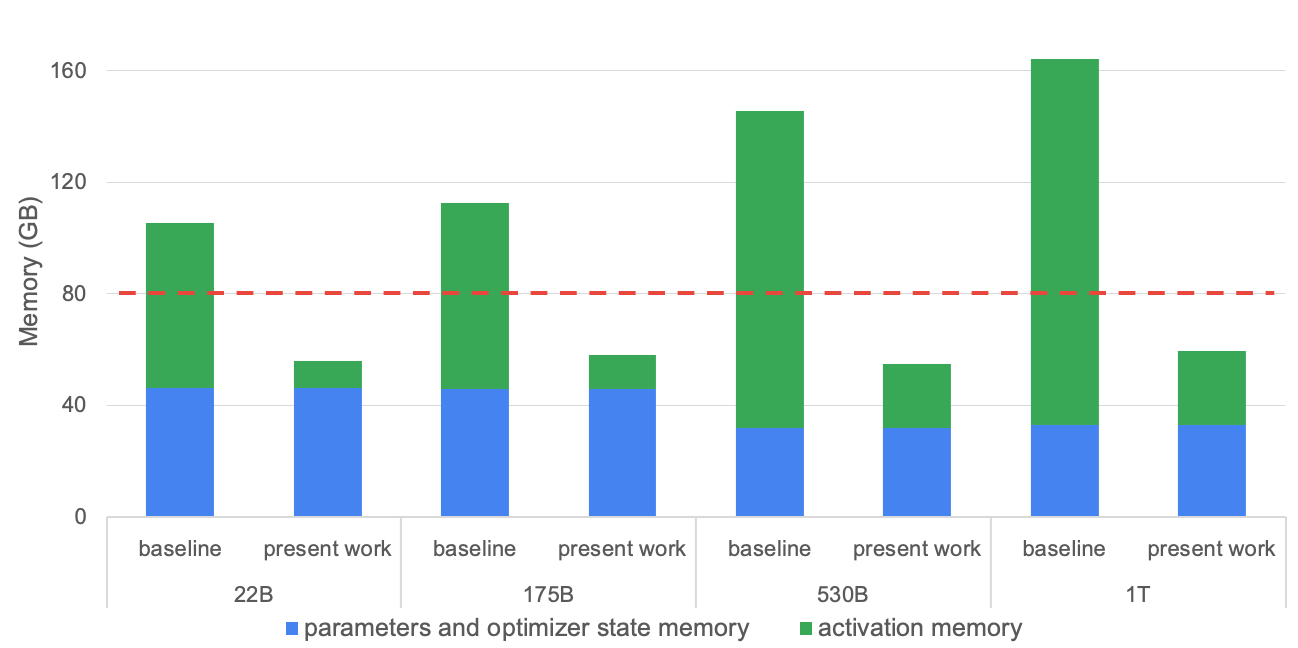}
  \caption{Parameters, optimizer state, and activations memory. The dashed red line represents the memory capacity of an NVIDIA A100 GPU. Present work reduces the activation memory required to fit the model. Details of the model configurations are provided in Table \ref{tab:configs}. }
  \label{fig:memory-model-activations}
\end{center}
\end{figure}

To quantify this, Figure \ref{fig:memory-model-activations} shows the memory required for four model configurations ranging from 22 billion parameters to 1 trillion parameters (details of the model configurations are provided in Table \ref{tab:configs}). It can be seen that for all these cases, the required memory for the baseline cases is above the 80GB memory provided by an NVIDIA A100 GPU. The standard approach to alleviate this memory pressure is to simply not store most of the activations and recompute them as necessary to calculate gradients during the backward pass \cite{activation-recomputation}. Unfortunately this method, usually called ``gradient checkpointing'' or ``activation recomputation'', incurs a steep penalty of reducing training efficiency. For transformer architectures, majority of prior work has checkpointed, or stored, the activations at transformer layer boundaries and recomputed the rest of the necessary activations in the backward pass. In this paper we refer to this method as ``full activation recomputation''. In our training runs, we observe $30-40\%$ execution time overhead when full activation recomputation is used.  

In this paper we present novel techniques that help alleviate the memory pressure of storing activations and thus reduce the need to recompute activations. These techniques are specific to the transformer architecture and are both simple to implement and have no, or very low, impact on compute efficiency. As we detail in Section~\ref{sec:related-work}, there are several other techniques to reduce the memory requirements of training large models, such as partitioning various data across the data parallel ranks or offloading data to CPU memory \cite{zero,zero-offload}. These techniques are complementary to the techniques presented here and could be additionally employed for even greater memory savings; however, in general these other techniques have both higher implementation cost and a larger impact on compute efficiency than the techniques presented in this paper. An analysis comparing these techniques to ours is outside the scope of this paper and left for future work.

We begin with a brief review of the transformer architecture and then build up an approximate formula for the memory required to store activations of a transformer model. Using this formula we can then study how different forms of model parallelism impact the activation memory requirements. We introduce sequence parallelism alongside tensor parallelism to prevent redundant storage of activations in regions that are not conducive to standard tensor parallelism. We then show that by being selective in what activations are saved and what are recomputed we can eliminate much of the cost of recomputation while using only a fraction of the memory when no recomputation is used. Finally, we present several experiments that measure the improvements these techniques make to both individual components of training as well as the full training throughput.

\section {Related Work}
\label{sec:related-work}

Model parallelism enables training very large models across multiple GPUs. Model parameters along with the associated optimizer states of these models require a huge amount of memory and do not fit on a single GPU. Even if we are able to fit the model in a single GPU (e.g., by swapping parameters between host and device memory \cite{zero-offload}), the high number of compute operations required can result in unrealistically long training times. This calls for parallelism. Two forms of model parallelism are commonly used to distribute the model parameters across GPUs: 1) tensor parallelism where parameters of each layer are distributed across many devices \cite{mesh-tensorflow, megatron, GSPMD}, and 2) pipeline parallelism where the model is split along the layer dimension of the network \cite{gpipe,terapipe,pipedream}. Some recent approaches combine both types of model parallelism to enable training large models up to 1T parameters \cite{megatron-pipeline}.

An alternative to model parallelism is to combine a number of training techniques along with data parallelism to enable large scale model training~\cite{zero, zero-offload, zero-infinity, deepspeed}. This approach is based on sharding the optimizer states, gradients, and parameters across data-parallel ranks. Also, a recent extension \cite{zero-infinity} uses CPU off-loading techniques to enable multi-trillion parameter model training on a small number of GPUs. Compared to model parallelism, these techniques, which are based on data parallelism, are less efficient and do not scale well to a large numbers of GPUs \cite{megatron-pipeline} and are thus a better fit for finetuning models in resource-constrained environments. This paper focuses only on model parallelism optimizations. An analysis comparing these techniques to ours is outside the scope of this paper.

% Covered in Introduction
%Along with optimizer states, parameters, and gradients, the activations stored during the forward propagation take up a significant portion of the memory. Techniques like activation checkpointing have been used to minimize activation memory at the expense of extra compute \cite{activation-recomputation}. Prior this work, activation checkpointing was commonly performed at the granularity of a transformer layer (referred to in this paper as full activation recomputation). Although performing activation recomputation at transformer layer granularity significantly reduces the activation memory requirement ($> 20X$), it adds $30-40\%$ overhead to execution time. In this paper, we present an alternative checkpoint granularity which has a better trade off between required activation memory and compute overhead. 

In addition, tensor parallelism as introduced in Megatron-LM \cite{megatron} helps to reduce the activation memory to some extent. In this approach, there are parts of the transformer where activations are not split across tensor parallel ranks, adding to activation memory overhead. Sequence parallelism as suggested in \cite{sequence-parallelism} where activations are partitioned along sequence dimensions throughout the network can alleviate this problem. However, their approach, similar to data parallelism, requires the parameters and optimizer state to be replicated on all of the devices which makes it not suitable for large model training. Sagemaker \cite{sagemaker} and GSPMD \cite{GSPMD} propose memory efficient versions of tensor parallelism which splits the activations across the devices along the hidden dimension throughout the network. The main drawback of these approaches is that they encompass multi-device layer normalization which is very compute/communication inefficient. Along with layer-norm communication, Sagemaker \cite{sagemaker} does four reduce-scatter communications per transformer layer in comparison to just two all-reduce communications in the Megatron-LM approach. %Although GSPMD \cite{GSPMD} uses similar communication pattern as ours, it does not get rid of inefficient distributed layernorm.
In this paper, we present a new technique which leverages the advantages of both tensor parallelism and sequence parallelism without any of the previous approaches' shortcomings. In other words, our technique, which mixes both tensor and sequence parallelism, reduces the activation memory significantly without any additional compute, communication, or memory overhead.

\section{Transformer Architecture}
\label{sec:transformer-arch}

\begin{figure}
\begin{center}
  \includegraphics[scale=0.25]{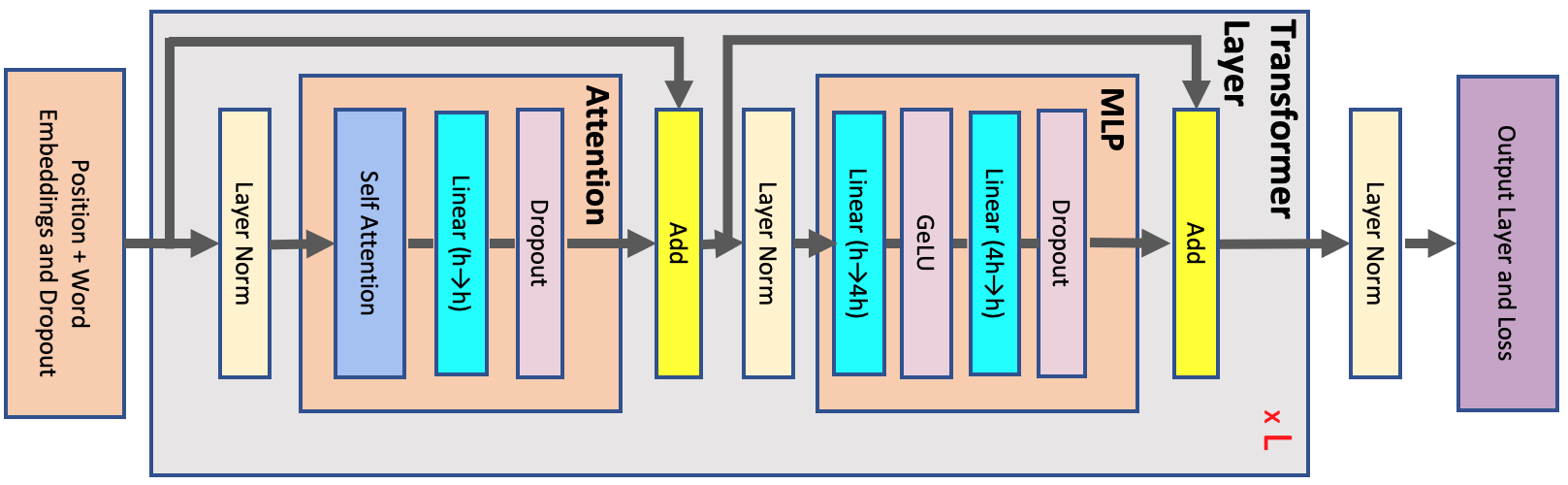}
  \caption{Transformer Architecture.  Each gray block represents a single transformer layer that is replicated $L$ times.}
  \label{fig:transformer-general}
\end{center}
\end{figure}

In this work, we consider a single stack transformer encoder or decoder with $L$ layers as shown in Figure \ref{fig:transformer-general}. At the start of the network, the input tokens are fed into a word embedding table with size $v \times h$ and the token embeddings are combined with learned positional embeddings with size $s \times h$ where $s$ is the sequence length, $h$ is the hidden dimension, and $v$ is the vocabulary size. The output of the embedding layer, which is the input to the transformer block, is a 3-D tensor of size $s \times b \times h$ where $b$ is the microbatch size. Each transformer layer consists of a self-attention block with $a$ attention heads followed by a multi-layer perceptron (MLP) with two layers which increase the hidden size to $4h$ and then reduce it back to $h$. Input to and output from each transformer layer have the same size $s \times b \times h$. The output from the last transformer layer is projected back into the vocabulary dimension to calculate the cross-entropy loss. We assume that word embedding and output layer weights are shared. Variable names are listed in Table~\ref{tab:variables} for reference.

\begin{table}
    \centering
    \begin{tabular}{c|lc|l}
        $a$ & number of attention heads & $p$ & pipeline parallel size \\
        $b$ & microbatch size & $s$ & sequence length \\
        $h$ & hidden dimension size & $t$ & tensor parallel size \\
        $L$ & number of transformer layers & $v$ & vocabulary size 
    \end{tabular}
    \caption{Variable names.}
    \label{tab:variables}
\end{table}

\section{Activation Memory}
\label{sec:activation-memory}

In this section, we derive an approximate formula for the memory required to store activations in the forward pass of a single stack transformer model as shown in Figure \ref{fig:transformer-general}. Note that ``activations'' in this paper refers to any tensor that is created in the forward pass and is necessary for gradient computation during back-propagation. As a result, this excludes the main parameters of the model and optimizer state, but, for example, includes the mask used by the dropout operation.

In addition, we only consider the main contributors to the memory and ignore small buffers. For example, for a layer normalization block, the input to the layer as well as the input's mean and variance are required to calculate the gradients. The input contains $sbh$ elements whereas mean and variance have only $sb$ elements each. Since $h$ is large (of order of thousands), we have $2sb \ll sbh$. As a result it is a good approximation to only consider the memory required to store the input, i.e., we only include $sbh$, not $sbh + 2sb$. 

We also assume that the network and the activations are stored in a 16-bit floating point format and therefore each element requires 2 bytes for storage. The only exceptions are the dropout masks which only require a single byte per element. Note that all the reported sizes in this section are in bytes and not number of elements unless explicitly mentioned.

\subsection{Activations Memory Per Transformer Layer}
\label{sec:act-mem-per-layer}

\begin{figure}
\begin{center}
  \includegraphics[scale=0.11]{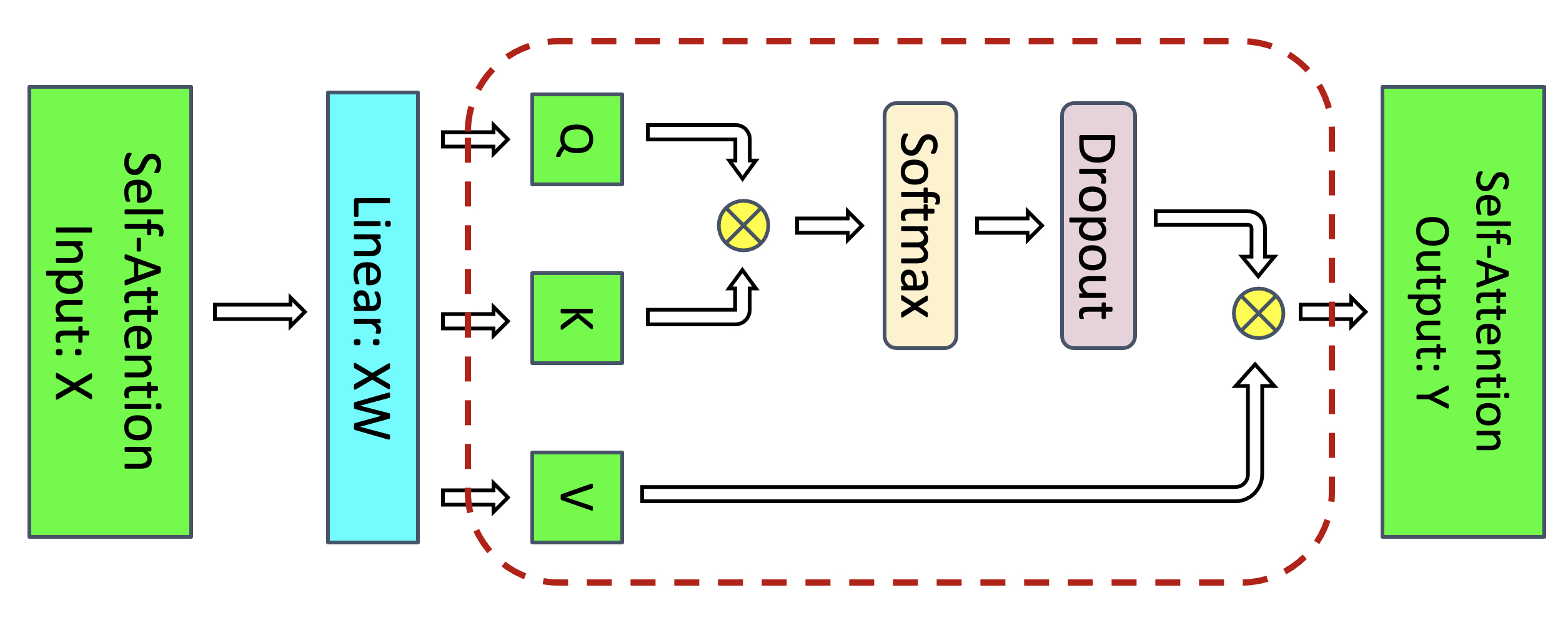}
  \caption{Self-attention block. The red dashed line shows the regions to which selective activation recomputation is applied (see Section \ref{sec:activation-recomputation} for more details on selective activation recomputation).}
  \label{fig:self-attention}
\end{center}
\end{figure}

As shown in Figure \ref{fig:transformer-general}, each transformer layer consists of an attention and an MLP block connected with two layer-norms. Below, we derive the memory required to store activations for each of these elements:

{\bf Attention block:} which includes self attention followed by a linear projection and an attention dropout. The linear projection stores its input activations with size $2sbh$ and the attention dropout requires a mask with size $sbh$. The self attention shown in Figure \ref{fig:self-attention} consists of several elements:
\begin{itemize}
    \item {\bf Query ($Q$), Key ($K$), and Value ($V$) matrix multiplies:} We only need to store their shared input with size $2sbh$.
    \item {\bf $QK^T$ matrix multiply:} It requires storage of both $Q$ and $K$ with total size $4sbh$.
    \item {\bf Softmax:} Softmax output with size $2as^2b$ is required for back-propagation.
    \item {\bf Softmax dropout:} Only a mask with size $as^2b$ is needed.
    \item {\bf Attention over Values ($V$):} We need to store the dropout output ($2as^2b$) and the Values ($2sbh$) and therefore need $2as^2b+2sbh$ of storage.
\end{itemize}
Summing the above values, in total, the attention block requires $11sbh + 5as^2b$ bytes of storage.

{\bf MLP:} The two linear layers  store their inputs with size $2sbh$ and $8sbh$. The GeLU non-linearity also needs its input with size $8sbh$ for back-propagation. Finally, dropout stores its mask with size $sbh$. In total, MLP block requires $19sbh$ bytes of storage.

{\bf Layer norm:} Each layer norm stores its input with size $2sbh$ and therefore in total, we will need $4sbh$ of storage.

Summing the memory required for attention, MLP, and the layer-norms, the memory required to store the activations for a single layer of a transformer network is:
\begin{equation}
    \text{Activations memory per layer = } sbh \left(34 + 5 \frac{as}{h}\right).
    \label{eq:memory-no-parallel}
\end{equation}
The above equation is for the case that no form of model parallelism is applied.

\subsection{Model Parallelism}
\label{sec:model-parallel}

In this section we start by quantifying the effect of tensor parallelism on the required activation memory per layer. We then introduce a novel method called sequence parallelism that further reduces per layer memory required by activations. At the end of this section, we also discuss the effect of pipeline parallelism on activations memory and derive a formula for the total memory required by activations.

\subsubsection{Tensor Parallelism}
\label{sec:tensor-parallel}

\begin{figure}[h!]
\begin{center}
  \includegraphics[scale=0.25]{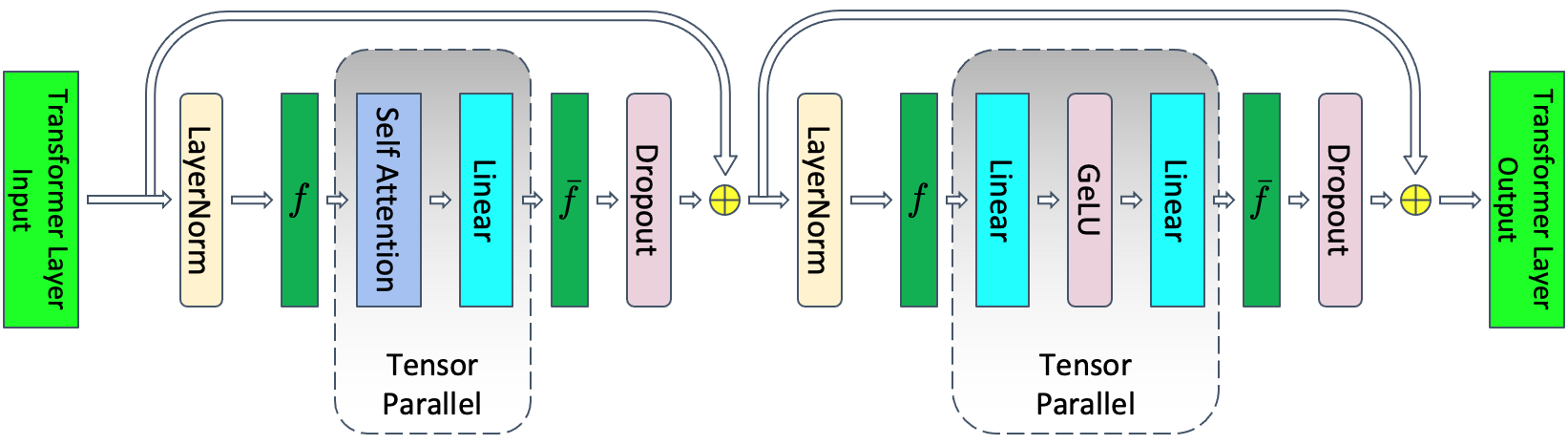}
  \caption{Transformer layer with tensor parallelism. $f$ and $\bar{f}$ are conjugate. $f$ is no operation in the forward pass and all-reduce in the backward pass. $\bar{f}$ is all-reduce in the forward pass and no operation in the backward pass.}
  \label{fig:transformer-tensor-parallel}
\end{center}
\end{figure}

We use tensor parallelism developed by Shoeybi, et.al.\cite{megatron} and parallelize the attention as well as MLP blocks as shown in Figure \ref{fig:transformer-tensor-parallel}. This form of parallelism introduces two additional communication operations $f$ and $\bar{f}$. For more details, please see the paper\cite{megatron}. 

Not only does tensor parallelism parallelize model parameters and optimizer states inside the attention and MLP blocks, but it also parallelizes the activations inside those blocks. Note that the input activations to these blocks (for example input to the $Q$, $K$, and $V$ matrix multiplies or input to the $h \rightarrow 4h$ linear layer) are not parallelized, and only activations within each block are divided across the tensor parallel group\footnote{The tensor parallel group is the group of accelerators that participate in tensor parallelism}. Assuming $t$-way tensor parallelism, the per-layer memory required to store the activations reduces from Equation \ref{eq:memory-no-parallel} to:
\begin{equation}
    \text{Activations memory per layer = } sbh \left(10 + \frac{24}{t} + 5 \frac{as}{ht}\right).
    \label{eq:memory-tensor-parallel}
\end{equation}

\subsubsection{Sequence Parallelism}
\label{sec:seq-parallel}

Tensor parallelism, as shown in Figure \ref{fig:transformer-tensor-parallel}, parallelizes the parts of the transformer layer that take the most time during training and as a result, it is computationally efficient. However, it leaves the layer-norms as well as the dropouts after attention and MLP blocks intact and as a result, they are replicated across the tensor parallel group. These elements do not require a lot of compute but demand a considerable amount of activation memory. Quantitatively, the $10sbh$ part of Equation \ref{eq:memory-tensor-parallel} is due to these replicated operations and as a result they are not divided by the tensor parallel size $t$.

We notice that in the non-tensor parallel regions of a transformer layer, the operations are independent along the sequence dimension. This characteristic allows us to partition these regions along the sequence dimension $s$. Partitioning along the sequence dimension reduces the memory required for the activations. This extra level of parallelism introduces new communication collectives before $f$ and after $\bar{f}$ which will act as converters between sequence and tensor parallel regions. For example, in the forward pass, we need an extra all-gather before the operator $f$ in Figure \ref{fig:transformer-tensor-parallel}. These extra communications introduce overhead and will slow down the training. 

\begin{figure}[h!]
\begin{center}
  \includegraphics[scale=0.25]{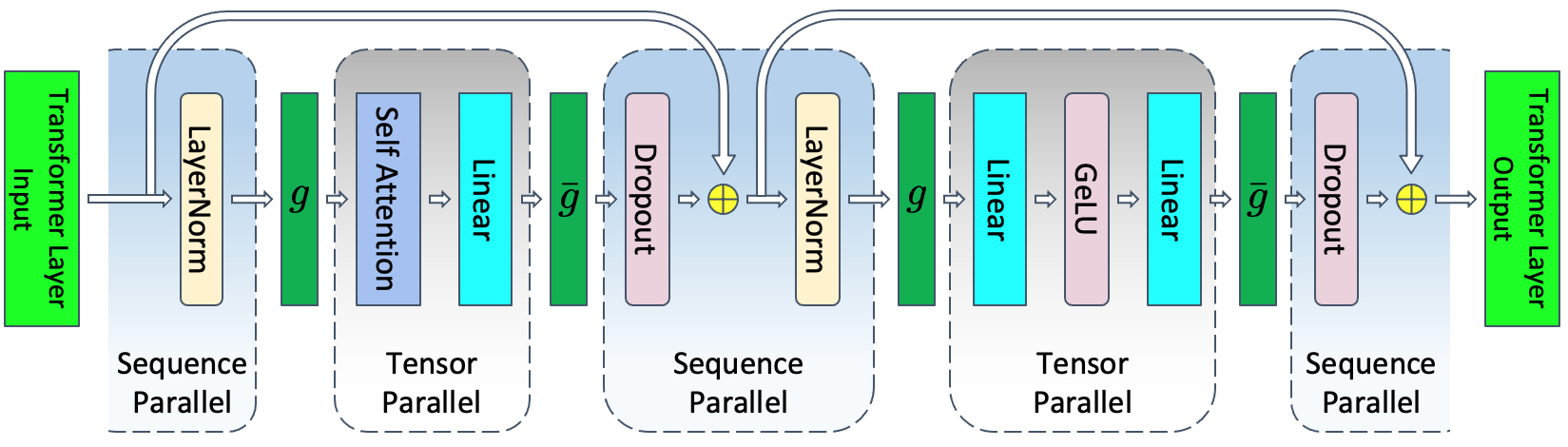}
  \caption{Transformer layer with tensor and sequence parallelism. $g$ and $\bar{g}$ are conjugate. $g$ is all-gather in the forward pass and reduce-scatter in the backward pass. $\bar{g}$ is reduce-scatter in forward pass and all-gather in backward pass.}
  \label{fig:transformer-tensor-sequence-parallel}
\end{center}
\end{figure}

To avoid these extra communications, we combine these operations with the $f$ and $\bar{f}$ operators and introduce new operations $g$ and $\bar{g}$ as shown in Figure \ref{fig:transformer-tensor-sequence-parallel}. As it can be seen, $g$ and $\bar{g}$ are the converters between sequence and tensor parallel regions. We derive these operations in the remainder of this section.

\begin{figure}
\begin{center}
  \includegraphics[scale=0.25]{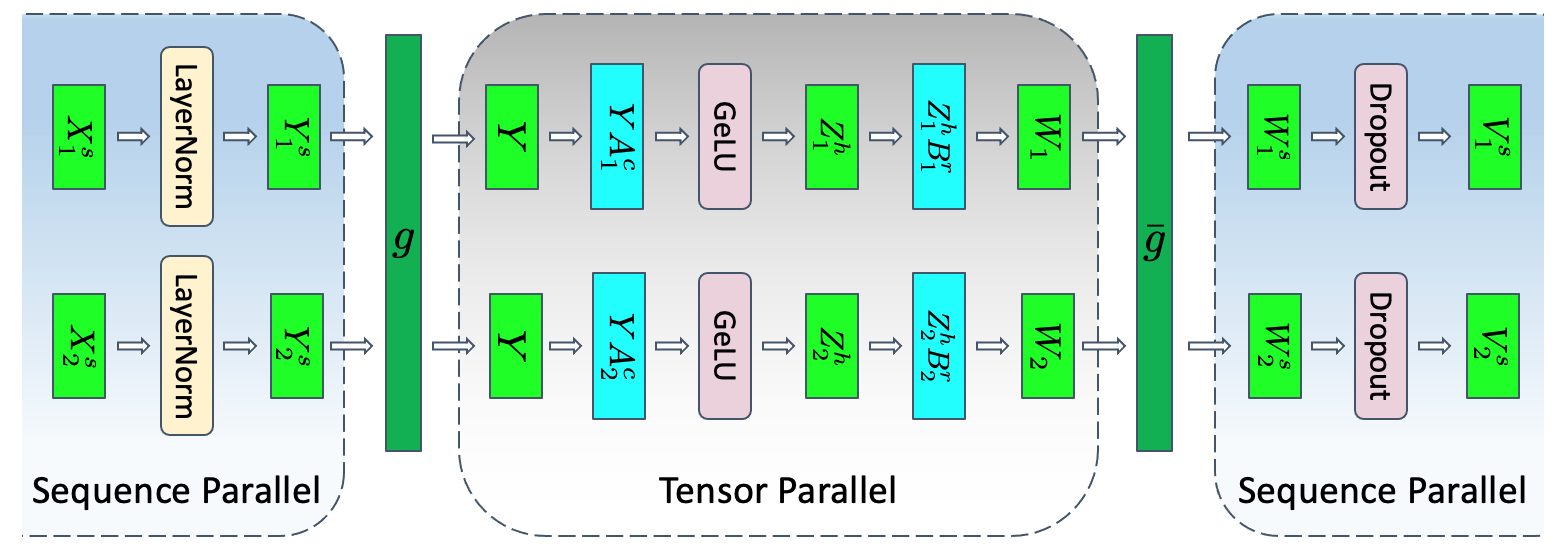}
  \caption{MLP layer with tensor and sequence parallelism. $g$ and $\bar{g}$ are conjugate. $g$ is all-gather in forward pass and reduce-scatter in backward pass. $\bar{g}$ is reduce-scatter in forward pass and all-gather in backward pass.}
  \label{fig:mlp-tensor-sequence-parallel}
\end{center}
\end{figure}

We detail $g$ and $\bar{g}$'s derivation using the MLP block. In the non-parallel form, as shown in Figure \ref{fig:transformer-general}, the layer-norm followed by the MLP block can be formulated as:
\begin{align*}
    Y &= \text{LayerNorm}(X), \\
    Z &= \text{GeLU}(YA), \\
    W &= ZB, \\
    V &= \text{Dropout}(W),
\end{align*}
where $X$ is input to the layer-norm with size $s\times b \times h$ and $A$ and $B$ are the weight matrices of the linear layers with size $h\times 4h$ and $4h\times h$, respectively. The combined tensor and sequence parallel form of the above operations is shown in Figure \ref{fig:mlp-tensor-sequence-parallel}. The subscripts represent splitting among accelerators and superscripts depict the dimension along which the splitting is done. For example, $X_1^s$ is the first accelerator's part of $X$ that is split along the $s$ dimension (sequence dimension) while $Z_2^h$  is the second accelerator's part of $Z$ that is split along the $h$ dimension (hidden dimension).

The input to the layer-norm is parallelized along the sequence dimension $X=[X_1^s, X_2^s]$. As a result, the output of the layer-norm will also be parallel along the sequence dimension $Y = [Y_1^s, Y_2^s]$. The linear layer with GeLU non-linearity requires the entire input $Y$ and therefore we need to perform an all-gather. This implies that $g$ is an all-gather operation along the sequence dimension in the forward pass. By splitting $A$ along its columns ($A_1^c$ and $A_2^c$) and $B$ along its rows ($B_1^r$ and $B_2^r$), we avoid communications (for more details please see \cite{megatron}) and arrive at $W_1$ and $W_2$. These two tensors are not parallel anymore and need to be summed as $W = W_1 + W_2$ before they are fed into the dropout layer. However, dropout needs its input to be parallel in the sequence dimension $s$. Instead of summing and then parallelizing in the sequence dimension, we combine these two operations into a reduce-scatter operation. As a result, $\bar{g}$ can be a single reduce-scatter operation in the forward pass. Putting it all together, we arrive at:
\begin{align}
    \begin{split}
        [Y_1^s, Y_2^s] &= \text{LayerNorm}([X_1^s, X_2^s]), \\
        Y &= g(Y_1^s, Y_2^s), \\
        [Z_1^h, Z_2^h] &= [\text{GeLU}(YA_1^c),\ \text{GeLU}(YA_2^c)], \\
        W_1 &= Z_1^hB_1^r \ \  \text{and} \ \  W_2=Z_2^hB_2^r, \\
        [W_1^s, W_2^s] &= \bar{g}(W_1, W_2), \\
        [V_1^s, V_2^s] &= [\text{Dropout}(W_1^s),\ \text{Dropout}(W_2^s)].
    \end{split}
    \label{eq:mlp-split}
\end{align}
If we follow a similar break-down for the backward pass, we find that $g$ and $\bar{g}$ are conjugate of each other. $g$ is an all-gather in the forward pass and a reduce-scatter in the backward pass, and $\bar{g}$ is a reduce-scatter in the forward pass and an all-gather in the backward pass. A similar breakdown done for the layer-norm followed by the attention part of the transformer layer arrives at Figure \ref{fig:transformer-tensor-sequence-parallel}.

Tensor parallelism requires four all-reduces in a single forward and backward pass whereas tensor together with sequence parallelism requires four all-gathers and four reduce-scatters in a single forward and backward pass. At the first look, it seems that tensor with sequence parallelism requires more communications compared to tensor parallelism. However, we note that a ring all-reduce is composed of two steps: a reduce-scatter followed by an all-gather. As a result, the communication bandwidth used for tensor parallelism and tensor together with sequence parallelism are the same. Therefore, sequence parallelism does not introduce any communication overhead.

From Equation \ref{eq:mlp-split}, sequence parallelism along with tensor parallelism divides all the activations required for the backward pass along the parallel dimension except for the tensor $Y$ that is required for the first linear operation. To alleviate this issue, we do not store the full tensor $Y$ for the backward pass. Instead, we store only the $Y_i^s$ part on the $i$th tensor parallel rank and perform an extra all-gather in the backward pass. To eliminate the latency introduced by this extra all-gather, we overlap this communication with the computation required to calculate gradients with respect to $Y$, and as a result, we reduce the overhead. 

Using sequence parallelism along with tensor parallelism, the memory required to store the activations per transformer layer reduces from Equation \ref{eq:memory-tensor-parallel} to:
\begin{equation}
     \text{Activations memory per layer = }  sbh \left(\frac{10}{t} +\frac{24}{t} + 5 \frac{as}{ht}\right) = \frac{sbh}{t} \left(34 + 5 \frac{as}{h}\right).
    \label{eq:memory-tensor-sequence-parallel}
\end{equation}
The above equation is now Equation \ref{eq:memory-no-parallel} divided by the tensor parallel size. This means that using tensor and sequence parallelism, we can distribute activations among the tensor parallel group and reduce the required memory by tensor parallel size $t$.

\subsubsection{Pipeline Parallelism}
\label{sec:pipeline-parallel}

Pipeline parallelism simply divides the $L$ layers of the transformer into $L/p$ groups of layers where $p$ is the pipeline parallel size. However, pipeline parallelism does not uniformly divide the total memory required for activations by $p$. This is due to the overlapping that pipeline parallel schedules introduce to reduce the pipeline bubble~\cite{megatron-pipeline}.

To quantify this, we consider the 1F1B pipeline schedule developed in PipeDream~\cite{1f1b-pipedream}. Schedules that have a minimized pipeline bubble put the most memory pressure on the first stage of the pipeline (first stage of the pipeline refers to first group of $L/p$ layers which also includes the input embeddings). A visualization of activation memory as a function of pipeline stage is shown in Appendix \ref{sec:app-pipeline-memory-optimization}. To keep the pipeline pressurized and avoid extra idle time, the first stage must store activations for $p$ microbatches (for more details see Figure 4-top of \cite{megatron-pipeline}). Each stage contains $L/p$ layers so the first stage must store $p \times L/p = L$ layers worth of activations regardless of the pipeline parallel size $p$. Therefore, the total memory required to store activations in the first stage is:
\begin{equation}
     \text{Total activations memory = } \frac{sbhL}{t} \left(34 + 5 \frac{as}{h}\right).
    \label{eq:memory-full-parallel}
\end{equation}

For other pipeline schedules, the total memory required would be slightly different. For example, the interleaving schedule developed in Megatron-LM~\cite{megatron-pipeline} requires storing activations for $L(1+\frac{p-1}{pm})$ layers where $m$ is the number of interleaving stages. As a result, if the interleaving schedule is used, then the total activation memory should be scaled by $(1+\frac{p-1}{pm})$.

\subsection{Total Activations Memory}
\label{sec:total-act-mem}

The majority of the required activation memory is captured by Equation \ref{eq:memory-full-parallel}. However, this equation does not capture activation memory required for the input embeddings, the last layer-norm, and the output layer as shown in Figure \ref{fig:transformer-general}.

Position and word embeddings do not require any considerable activations to be stored for the backward pass. However, the dropout requires storage. The dropout in the embeddings layer is also parallelized along the sequence dimension. As a result, it will require $sbhp/t$ storage. Note that the factor $p$ comes from the pipeline parallelism and the fact that we need to store $p$ microbatches (see Section \ref{sec:pipeline-parallel}).

The layer-norm before the output layer also uses sequence parallelism and as a result requires $2sbh/t$ storage. The output layer projection into vocabulary dimension will require its input with size $2sbh/t$ to be stored. Finally, the cross entropy loss requires storing the logits which are calculated in 32-bit floating point and as a result will require $4sbv/t$ of storage. Note that since we only consider activations in the first stage of the pipeline, the above activations, i.e., $4sbh/t(1+v/h)$ in total, are only included for the case that there is no pipeline parallelism ($p=1$).

Adding the above memory, the extra memory due to the input embeddings, the last layer-norm, and the output layer is:
\begin{equation*}
     \frac{sbhL}{t} \left(\frac{p}{L} + \delta_{p=1}\frac{4}{L}\left(1+\frac{v}{h}\right)\right)
    \label{eq:memory-full-parallel-with-first-last}
\end{equation*}
where $\delta_{p=1}$ is $1$ for $p=1$ and $0$ otherwise. We note that compared to term $34+5\frac{as}{h}$ from Equation \ref{eq:memory-full-parallel}, both $p/L$ and $4/L(1+v/h)$ are negligible. For example, for a model with 22B parameters, these extra terms account for less than $0.01\%$ of the total activation memory requirements. As a result, Equation \ref{eq:memory-full-parallel} is a good approximation to the total required activations memory and we will use it in the rest of this paper.

\section{Selective Activation Recomputation}
\label{sec:activation-recomputation}

The total required activation memory from Equation \ref{eq:memory-full-parallel} can still be considerable for large models. Activation recomputation \cite{activation-recomputation} overcomes this memory limitation by storing (or "checkpointing") the input activations to a group of layers and recomputing other required activations using an extra forward pass during back-propagation (this is referred to in this paper as full activation recomputation). Assuming the groups contain only a single layer, and ignoring activations outside of the transformer layers, this method reduces the total required memory for activations to $2sbhL$. We note that this required memory can be further reduced to $2sbhL/t$ if we only store a portion of activations in each tensor parallel rank. However, this approach requires an extra all-gather per layer and will add communication overhead and, as a result, we do not consider this approach.

Compared to storing all activations (Equation \ref{eq:memory-full-parallel}), checkpointing all transformer layers significantly reduces the amount of memory required to train a model. % by $17/t + 5as/2ht$ times.
This reduction does come at the cost of the recomputation (an extra forward pass) which can introduce as much as $30-40\%$ computational time overhead. To balance the memory savings and computational overhead, it is ideal to only checkpoint enough activations to allow a given model-parallel configuration to train given the constraints of device memory. The memory savings provided by sequence parallelism allows many more configurations to train without recomputation than before, but the optimal model parallel configurations of large models still generally require some saving and recomputing of activations. A simple approach to choose the amount of activations that are stored vs recomputed is to only checkpoint some of the transformer layers and store all the activations of other layers. This approach does not scale very well to large models; for example, when training MT-NLG there are only three layers per device, limiting the granularity at which you can balance memory vs compute. Additionally, we note that not all activations require the same amount of operations to recompute so it is beneficial to be smarter in selecting which activations to store and which to recompute.

Instead of checkpointing and recomputing full transformer layers, we propose to checkpoint and recompute only parts of each transformer layer that take up a considerable amount of memory but are not computationally expensive to recompute, or selective activation recomputation. To this end, we note that the term $5as/h$ in Equation \ref{eq:memory-full-parallel} is due to the attention operations after the width of the network is increased by the linear layer calculating the Q, K, and V values; i.e., $QK^T$ matrix multiply, softmax, softmax dropout, and attention over $V$ as shown in Figure \ref{fig:self-attention}. These operations generally have large input sizes and thus large activations, however, the number of floating-point operations (FLOPs) per input element is very low. The rest of the transformer layer accounts for the 34 term in Equation~\ref{eq:memory-full-parallel}. Thus, for large models where $5as/h > 34$, if we checkpoint and recompute this part of the transformer layer, we store less than half of the activations and only have a modest cost to recompute those that aren't stored. 

\begin{table}
    \centering
    \begin{tabular}{c|c}
        Configuration & Activations Memory Per Transformer Layer \\
        \hline
        \multirow{2}{*}{no parallelism} & \multirow{2}{*}{$sbh \left(34 + 5 \frac{as}{h}\right)$} \\ \\ \hline
        \multirow{2}{*}{tensor parallel (baseline)} & \multirow{2}{*}{$sbh \left(10 + \frac{24}{t} + 5 \frac{as}{ht}\right)$} \\ \\ \hline
        \multirow{2}{*}{tensor + sequence parallel} & \multirow{2}{*}{$sbh \left(\frac{34}{t} + 5 \frac{as}{ht}\right)$} \\ \\ \hline
        tensor parallel +  & \multirow{2}{*}{$sbh \left(10 + \frac{24}{t}\right)$} \\ 
        selective activation recomputation & \\ \hline
        tensor parallel + sequence parallel +  & \multirow{2}{*}{$sbh(\frac{34}{t})$} \\ 
        selective activation recomputation & \\\hline
        \multirow{2}{*}{full activation recomputation} & \multirow{2}{*}{$sbh(2)$} \\ &
    \end{tabular}
    \caption{Activations memory, in bytes, per transformer layer for different techniques. }
    \label{tab:memory-eqns}
\end{table}

To quantify this, let's consider GPT-3 \cite{gpt-3} and MT-NLG \cite{mt-nlg} models, some of the largest models that have been trained so far. For GPT-3, $a=96$, $s=2048$, and $h=12288$ and as a result $5as/h=80$. For MT-NLG, $a=128$, $s=2048$, and $h=20480$ so $5as/h=64$. Comparing these numbers to 34, which is the factor for the rest of the layer, we can see these activations account for a large portion of the total activations. Thus, by using selective activation recomputation we can save $70\%$ and $65\%$ of the required memory for activations for the GPT-3 and MT-NLG models, respectively. The recomputation of these activations introduces only $2.7\%$ and $1.6\%$ FLOPs overhead for these two models. For more details on the FLOPs calculations see Appendix \ref{sec:app-FLOPs}.

Using this form of selective activation recomputation, the memory needed to store activation decreases from Equation \ref{eq:memory-full-parallel} to:
\begin{equation}
     \text{Total required memory = } 34\frac{sbhL}{t}.
    \label{eq:memory-full-parallel-partial}
\end{equation}

The above equation shows that using selective activation recomputation allows the required activation memory to scale linearly with sequence length and be independent of the number of attention heads. As was discussed in Section \ref{sec:pipeline-parallel}, in the case of an interleaved pipeline schedule, the above equation needs to be multiplied by $(1+\frac{p-1}{pm})$. 

When using pipeline parallelism, as discussed in Section~\ref{sec:pipeline-parallel}, even though a given device only has $L/p$ layers, the first stage must still store an equivalent of $L$ layers of activations since it must store activations for $p$ microbatches to keep the pipeline pressurized. An additional technique that can be employed to reduce the recomputation cost in this case is to store all the activations for as many microbatches as possible given available device memory, and do full or selective recomputation of the rest. In practice we find that after applying sequence parallelism and selective activation recomputation the recomputation overhead is small enough that this additional technique provides very modest improvement. This technique is described in more detail and analyzed in Appendix~\ref{sec:app-mb-recompute}.

\section{Evaluations}
\label{sec:evaluations}

In this section, we evaluate the impact of our proposed approach on both memory usage as well as execution speed of training. Table~\ref{tab:configs} lists the model configurations used in the evaluations. We consider models up to one trillion parameters and for all these models, the tenor parallel size is set to 8. We use the interleaving schedule with three interleaving stages ($m=3$) for the 175B and 530B models. For all the cases, sequence length is set to $s=2048$ and vocabulary size is set to $v=51200$. We also note that no data parallelism is considered in these evaluations since our approach is independent of data parallelism. As a result, the batch sizes used in our analysis are much lower than the ones used for the end-to-end training. All of our results are run with mixed precision on the Selene supercomputer~\cite{selene}. Each cluster node has 8 NVIDIA 80GB A100 GPUs~\cite{a100} connected to each other by NVLink and NVSwitch~\cite{nvlink}. Each node has eight NVIDIA Mellanox 200Gbps HDR Infiniband HCAs for application communication.

\begin{table}
    \centering
    \begin{tabular}{c|c|c|c|c|c|c|c|c}
        Model & Attention & Hidden & \multirow{3}{*}{Layers} & Tensor        & Pipeline      & Number & Global      & Micro   \\
                           Size   & Heads     & Size   &                         & Parallel & Parallel & of & Batch & Batch \\
                           & & & & Size & Size & GPUs & Size & Size \\
        \hline
        22B  & 64  & 6144  & 48  & 8 & 1 & 8 & 4   & 4  \\
        175B (GPT-3) & 96  & 12288 & 96  & 8 & 8 & 64  & 64  & 1\\
        530B (MT-NLG) & 128 & 20480 & 105 & 8 & 35 & 280 & 280 & 1\\
        1T   & 160 & 25600 & 128 & 8 & 64 & 512 & 512  & 1 \\
    \end{tabular}
    \caption{Model configurations used during evaluation. Note that no data parallelism is used in our evaluations and as a result, the batch sizes as well as total number of GPUs are set to a value much lower than the ones in the end-to-end training.}
    \label{tab:configs}
\end{table}

\subsection{Memory Usage}
\label{sec:memory-usage}

\begin{figure}
\begin{center}
  \includegraphics[scale=0.27]{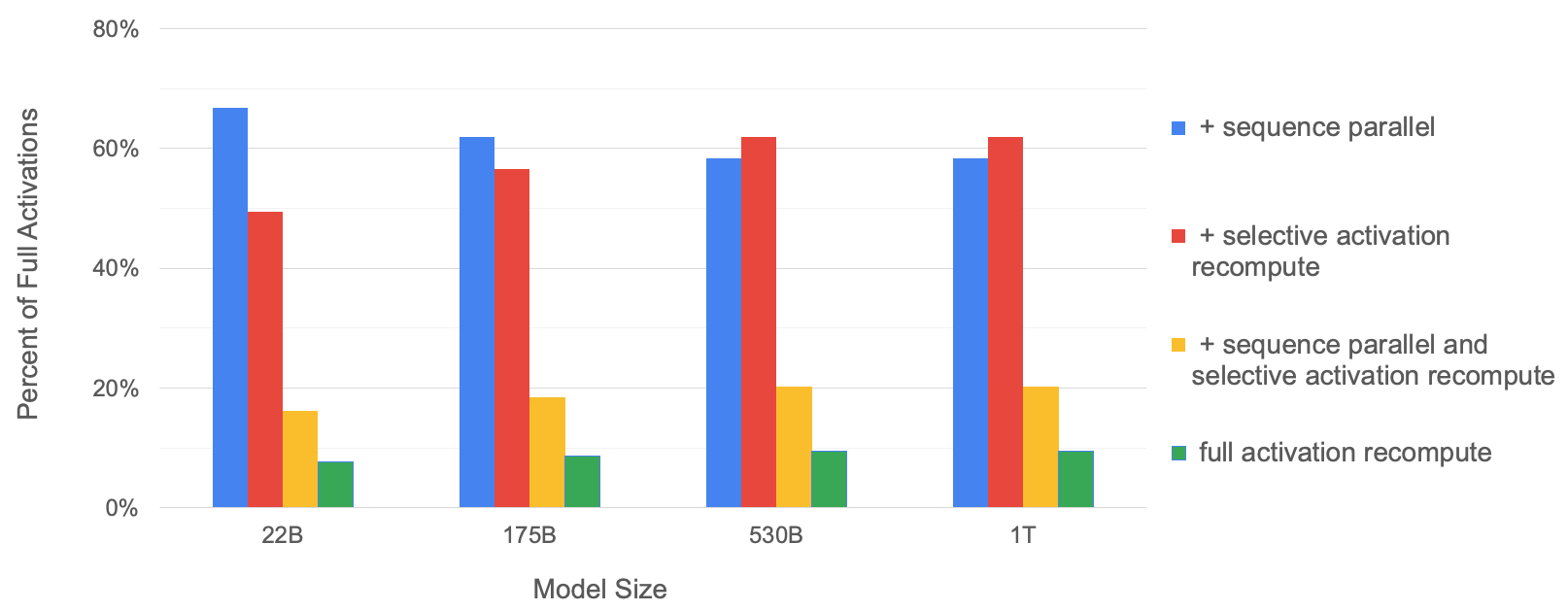}
  \caption{Percentage of required memory compared to the tensor-level parallel baseline. As the model size increases, both sequence parallelism and selective activation recomputation have similar memory savings and together they reduce the memory required by $\sim 5\times$.}
  \label{fig:percentage-full-activations}
\end{center}
\end{figure}

Table \ref{tab:memory-eqns} summarizes the required memory for different technique discussed in this paper. To quantify this, Figure~\ref{fig:percentage-full-activations} shows the activation memory used by different techniques as a percentage of the memory needed to keep all activations split across the tensor parallel ranks, i.e., Equation \ref{eq:memory-tensor-parallel}. Individually, both techniques cut the memory requirement nearly in half, and combined provide a 5x reduction bringing the memory requirements to under 20\%. This is only $\sim 2\times$ of the full activation recomputation which is at $10\%$ of the baseline. Without the memory savings provided by sequence parallelism and selective recompute together, none of the these models will fit into memory. Note that all of these results include the memory optimization described in Appendiex~\ref{sec:app-pipeline-memory-optimization}.

\subsection{Execution Time per Layer}
\label{sec:executation-time}

Table~\ref{tab:layer_times} shows the time to execute the forward and backward passes of one transformer layer of the 22B model for various experiments\footnote{These experiments were done on the 22B model with just one layer so that they would fit into device memory without any recomputation to obtain a baseline.}. The first two rows show that sequence parallelism provides a modest improvement to the time it takes to complete one transformer layer, reducing the forward time from 7.7ms to 7.2ms, a 6\% speedup. This improvement comes from the layer-norm and dropout layers being performed on $1/t$ of the data. We also found that even though the amount of data moved is the same, the execution of reduce-scatter and all-gather combined is slower than an all-reduce alone, reducing the improvement from sequence parallelism. Note that this speedup is an additional benefit to the primary advantage of using sequence parallelism, which is the memory savings that allow for less recomputation of activations.

The next two rows in Table~\ref{tab:layer_times} show that if we are selective in what operations are recomputed, which we can be in more configurations thanks to sequence parallelism, we can significantly reduce the overhead of recomputation in the backward pass. The overhead of selective recomputation is 1.3ms, or 11\% of the 11.9ms baseline, vs 7.6ms or 64\% overhead for recomputing the full layer. For the combined forward and backward time, the overhead is 7\% vs 39\%. Note that the overhead of 39\% for recomputing the full layer (as opposed to the expected 33\%) is due to an optimization in the backward pass where we overlap all-reduce communication with the linear weight's gradient computation. As we see later, this benefit increases with model size. The bottom row in Table~\ref{tab:layer_times} shows the combined benefit of selective recomputation and sequence parallelism. When the two techniques are used together, the overhead drops to just $4\%$.

\begin{table}
    \centering
    \begin{tabular}{r|c|c|c|c}
         Experiment & Forward (ms) & Backward (ms) & Combined (ms) & Overhead (\%) \\
         \hline
         Baseline no recompute & 7.7 & 11.9 & 19.6 & --\\
         Sequence Parallelism  & 7.2 & 11.8 & 19.0 & $-3\%$\\
         \hline
         Baseline with recompute & 7.7 & 19.5 & 27.2 & $39\%$ \\
         Selective Recompute & 7.7 & 13.2 & 20.9 & $7\%$ \\
         \hline
         Selective + Sequence & 7.2 & 13.1 & 20.3 & $4\%$ \\
    \end{tabular}
    \caption{Time to complete the forward and backward pass of a single transformer layer of the 22B model.}
    \label{tab:layer_times}
\end{table}

\begin{figure}
    \centering
    \includegraphics[width=\columnwidth]{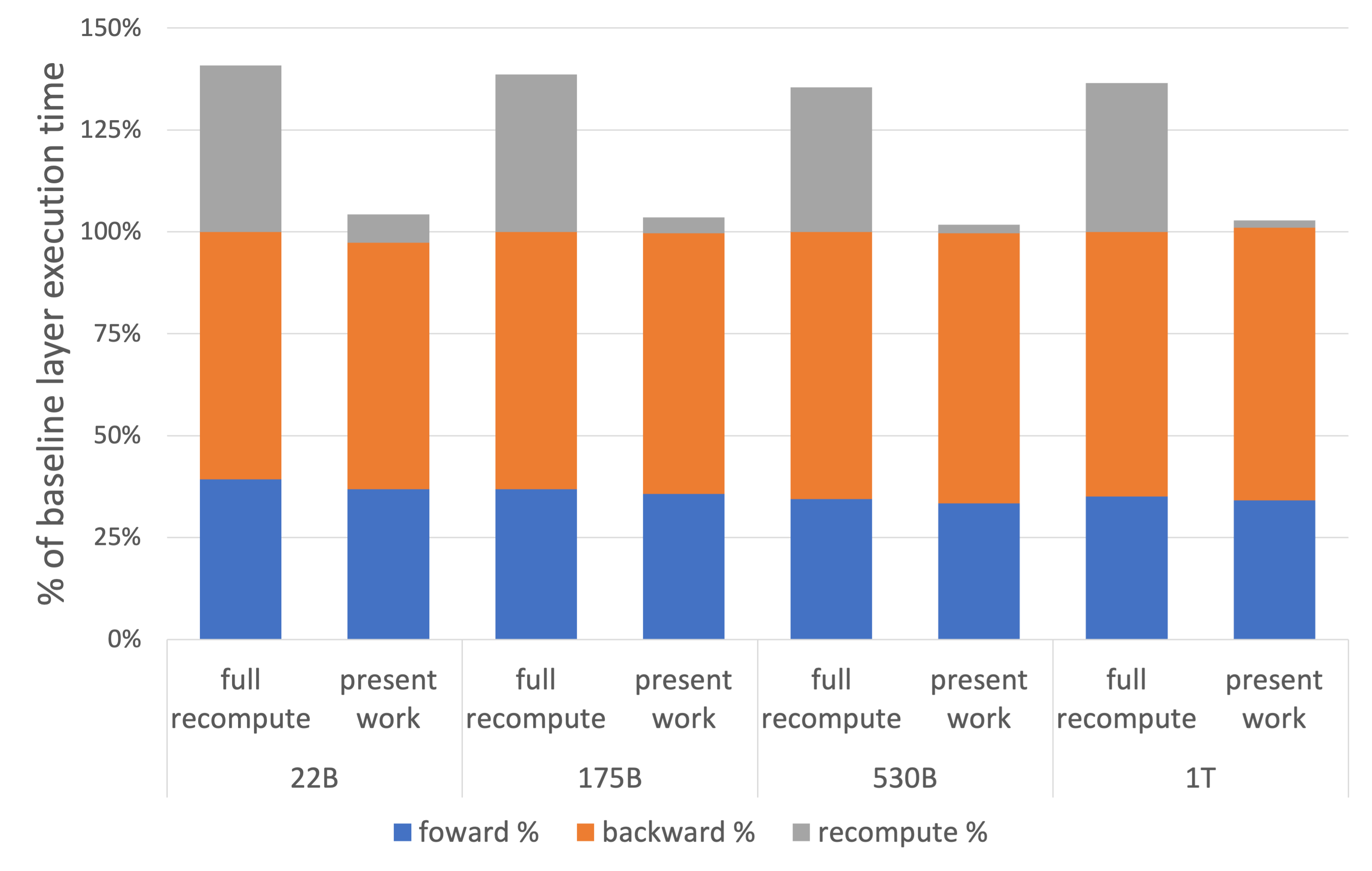}
    \caption{Per layer breakdown of forward, backward, and recompute times. Baseline is the case with no recomputation and no sequence parallelism. Present work includes both sequence parallelism and selective activation recomputation.}
    \label{fig:execution_time}
\end{figure}

Figure~\ref{fig:execution_time} shows this same break down for all of our test cases. We see that as the model size grows, the reduction in overhead also increases. For the 530B and 1T cases, the overhead is just $2\%$, compared to $36\%$ overhead for full recompute.

\subsection{End-to-End Iteration Time}
\label{sec:e2e-iteration-time}

Table~\ref{tab:iteration_time} lists full end-to-end iteration time for each of the four configurations listed in Table \ref{tab:configs}. We find that for all of the tested configurations, the techniques presented in the paper provide between $29.0\%$ and $32.1\%$ improvement in the throughput over performing full recomputation without sequence parallelism. These savings will be directly translated into shorter training times.  

\begin{table}
    \centering
    \begin{tabular}{c|c|c|c|c|c}
         \multirow{2}{*}{Model Size} & \multicolumn{2}{c|}{Iteration Time (seconds)} &
         Throughput & Model FLOPs & Hardware FLOPs\\
         \cline{2-3}
          & Full Recompute & Present Work & Increase & Utilization & Utilization\\
          \hline
          22B   & 1.42  & 1.10  & 29.0\% & 41.5\% & 43.7\% \\
          175B  & 18.13 & 13.75 & 31.8\% & 51.4\% & 52.8\% \\
          530B  & 49.05 & 37.83 & 29.7\% & 56.0\% & 57.0\% \\
          1T    & 94.42 & 71.49 & 32.1\% & 56.3\% & 57.0\% \\
    \end{tabular}
    \caption{End-to-end iteration time. Our approach results in throughput increase of around $30\%$.}
    \label{tab:iteration_time}
\end{table}

We define the model FLOPs utilization (MFU) and hardware FLOPs utilization (HFU) similar to Chowdhery, et al.~\cite{palm}. Model FLOPs are the floating point operations required to perform a single forward and backward pass (single iteration) regardless of the implementations and hardware limitations. As a result, model FLOPs are hardware and implementation independent and only depend on the underlying model. On the other hand, the hardware FLOPs represent the floating point operations that are actually performed on the hardware per iteration. Therefore, if an implementation requires activation recomputation (for example ours), then the hardware FLOPs are going to be larger than model FLOPs. We provide a tight lower bound formula for the model and hardware FLOPs in Appendix \ref{sec:app-FLOPs}. For our method, the hardware to model FLOPs ratio is approximately $1+s/6h$.

Subsequently, we define the model and hardware FLOPs per second as the model and hardware FLOPs divided by the iteration time, respectively. Using these definitions, the MFU and HFU are defined as model and hardware FLOPs per second divided by the accelerator theoretical peak FLOPs per second\footnote{For NVIDIA A100 GPUs, the peak theoretical FLOPs per second is 312 teraFLOPs/sec.}. Both the MFU and HFU are provided in Table~\ref{tab:iteration_time} for all four configurations. As the model size increases, we achieve better GPU utilization and for the one trillion parameter model, we reach a MFU and HFU of $56.3\%$ and $57.0\%$, respectively.

We should note that the results in Table~\ref{tab:iteration_time} do not use any data parallelism. Data parallelism introduces some overhead due to the gradient all-reduce required between the data parallel groups. However, for large transformer models, this overhead is not large. For example, if we scale the 530B model to 8-way data parallellism (2240 GPUs) while keeping batch size per model instance constant -- i.e., the batch size is also multiplied by the data parallel size -- the time per iteration increases slightly from $37.83$ seconds to $39.15$ seconds. This results in an MFU drop from $56.0\%$ to $54.2\%$ which is not substantial. We note that we do not use any overlapping of gradient all-reduces with back-propagation and an efficient overlap can almost entirely eliminate this increase in the iteration time.

\section{Conclusions and Future Work}
\label{sec:conclusions}

In this work, we presented two novel and straightforward techniques that reduce memory pressure from storing activations and thus reduce the need to recompute activations. We showed that using sequence parallelism with tensor parallelism can substantially reduce the required activation memory. In conjunction with selective activation recomputation, we showed that we can achieve a $5\times$ reduction in memory and recover over $90\%$ of the compute overhead introduced using full activation recomputation. 

In the future, we plan to further reduce the activation memory by resolving the issues arising from memory fragmentation for large microbatches and non-uniform memory allocation due to pipeline parallelism. Moreover, we plan to work on methods that can reduce the memory pressure on the first stage of the pipeline. 
%utilize the freed up memory to bring down the compute overhead even further by avoiding activation recomputation on a subset of layers.

\bibliographystyle{plain}
\bibliography{megatron-sp-sl}

\appendix

\section{FLOPs Calculation}
\label{sec:app-FLOPs}

For FLOPs calculations, we follow the derivation from Narayanan, et.al.\cite{megatron-pipeline} and only consider the matrix multiplications (GEMMs) which are the main contributors to the number of floating-point operations. For the attention block, the main contributors to floating-point operations are: key, query, and value transformation ($6Bsh^2$ operations), attention matrix computation ($2Bs^2h$ operations), attention over values ($2Bs^2h$ operations), and post-attention linear projection ($2Bsh^2$ operations) where $B$ is the microbatch size.

For the feed-forward network that increases the hidden size to $4h$ and then reduces it back to $h$,  we have $16Bsh^2$ floating-point operations. Summing these together, each transformer layer results in $24Bsh^2 + 4Bs^2h$ FLOPs for the forward pass. The other main contributor to the number of floating-point operations is the logits layer in the language model head, which transforms features of dimension $h$ to the vocabulary dimension $v$. The required FLOPs for this operation is $2Bshv$.

The backward pass requires double the number of FLOPs since we need to calculate the gradients with respect to both input and weight tensors. Summing all the contributions, the number of FLOPs required to do one forward and one backward pass (denoted by model FLOPs) is:
\begin{equation}
    \text{model FLOPs per iteration} = 72BLsh^2\left( 1 + \frac{s}{6h} + \frac{v}{12hL}\right).
    \label{eq:model-flops}
\end{equation}

Selective activation recomputation requires an additional forward pass attention matrix computation ($2Bs^2h$ operations) and attention over values ($2Bs^2h$ operations). Adding these required FLOPs to Equation \ref{eq:model-flops}, the total number of FLOPs we require per iteration (denoted by hardware FLOPs) is:
\begin{equation}
    \text{hardware FLOPs per iteration} = 72BLsh^2\left( 1 + \frac{s}{3h} + \frac{v}{12hL}\right).
    \label{eq:hardware-flops}
\end{equation}

One can see that in our approach, model FLOPs are very close to the hardware FLOPs. Assuming $3h\gg s$ and $12hL\gg v$ and only considering main terms, the ratio of the hardware to model FLOPs can be approximated as:
\begin{equation}
    \frac{\text{hardware FLOPs}}{\text{model FLOPs} } \approx 1 + \frac{s}{6h}.
    \label{eq:hardware-model-flops-ratio}
\end{equation}

%\begin{equation*}
%    \frac{\text{hardware FLOPs per iteration}}{\text{model FLOPs per iteration} } = \frac{1 + \frac{s}{3h} + \frac{v}{12hL}}{1 + \frac{s}{6h} + \frac{v}{12hL}}
%\end{equation*}

\section{Pipeline Parallelism Memory Optimization}
\label{sec:app-pipeline-memory-optimization}

\begin{figure}
\begin{center}
  \includegraphics[scale=0.5]{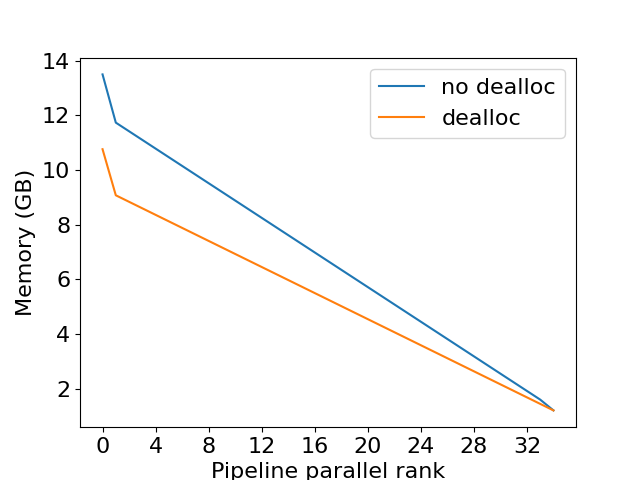}
  % \caption{Total activation memory used by each pipeline parallel rank, shown for both an un-optimized case (blue), and when using our baseline memory optimization that deallocates the output tensor of each pipeline rank (yellow). The model used here had tensor parallel size 8, pipeline parallel size 16, sequence length 2048, hidden size 32768, and micro batch size 8.}
  \caption{Activation memory of each pipeline parallel rank, shown for both an unoptimized case (blue), and with our memory optimization that deallocates the output tensor of each pipeline rank (yellow). The 530B model (see Table \ref{tab:configs}) is used for this experiment.}
  \label{fig:pipeline-parallel-eval}
\end{center}
\end{figure}

% Rather, activation memory decreases linearly from the first stage to the last stage, with the first stage containing additional activation memory for the embedding layer, as detailed in Section \ref{sec:total-act-mem}.
As mentioned in Section \ref{sec:pipeline-parallel}, efficient pipeline parallelism strategies do not uniformly divide the total memory across pipeline ranks. Figure \ref{fig:pipeline-parallel-eval} illustrates this memory pattern for the 530B model detailed in Table \ref{tab:configs}, showing the memory imbalance across pipeline ranks for both an unoptimized case, and when deallocating each microbatch's output tensor after its forward pass. This optimization relies on the fact that, after a microbatch's forward pass, the output tensor's data is redundant with the input data of the following stage, thereby making it safe to deallocate. We should note that in all other results shown in this paper, this output-tensor-deallocation optimization is used.

% In this figure, we can observe the linear memory decrease along pipeline ranks, and the additional $sbhp$ activation memory stored on the first pipeline rank for the embedding layer (see Section \ref{sec:total-act-mem}). We can also observe the activation memory flattening at zero on the higher ranks, due to ???
In this figure, we can observe the linear memory decrease along the pipeline ranks (directly observable for pipeline ranks 1 and above), along with the additional $sbhp$ memory spike on pipeline rank 0 due to the embedding layer's activation memory (see Section \ref{sec:total-act-mem}).

% In the case of applying the output-tensor-deallocation optimization, we save $sbhm$ memory per pipeline rank, where $m$ is the number of microbatches \textit{in flight} on each rank, peaking at $m = p$ on the first pipeline stage. Using the hyperparameters listed in Figure \ref{fig:pipeline-parallel-eval}, and this setup's 2 bytes per data element, the theoretical savings for this optimization on the first pipeline stage is $sbhp =$ 16 GB, which closely matches the difference between the lines shown in the figure for rank 0.
In the case of applying the output-tensor-deallocation optimization, we save $sbhr$ memory per pipeline rank, where $r$ is the number of microbatches \textit{in flight} on each rank, peaking at $r = p$ on the first pipeline stage. Using the hyperparameters for the 530B model (see Table \ref{tab:configs}), and this setup's 2 bytes per data element, the theoretical savings for this optimization on the first pipeline stage is $sbhp =$ 2.73 GB, which closely matches the difference between the lines shown in the figure for rank 0.

\section{Microbatch Level Activation Recomputation}
\label{sec:app-mb-recompute}

GPUs used in pipeline parallel model training store the input activations of layers until they are consumed at the gradient computation during back-propagation. As discussed in Section~\ref{sec:pipeline-parallel}, the first pipeline stage stores the most activations, an equivalent of storing activations for all of the transformer layers in the model. 

\begin{figure}[h!]
\begin{center}
  \includegraphics[width=\columnwidth]{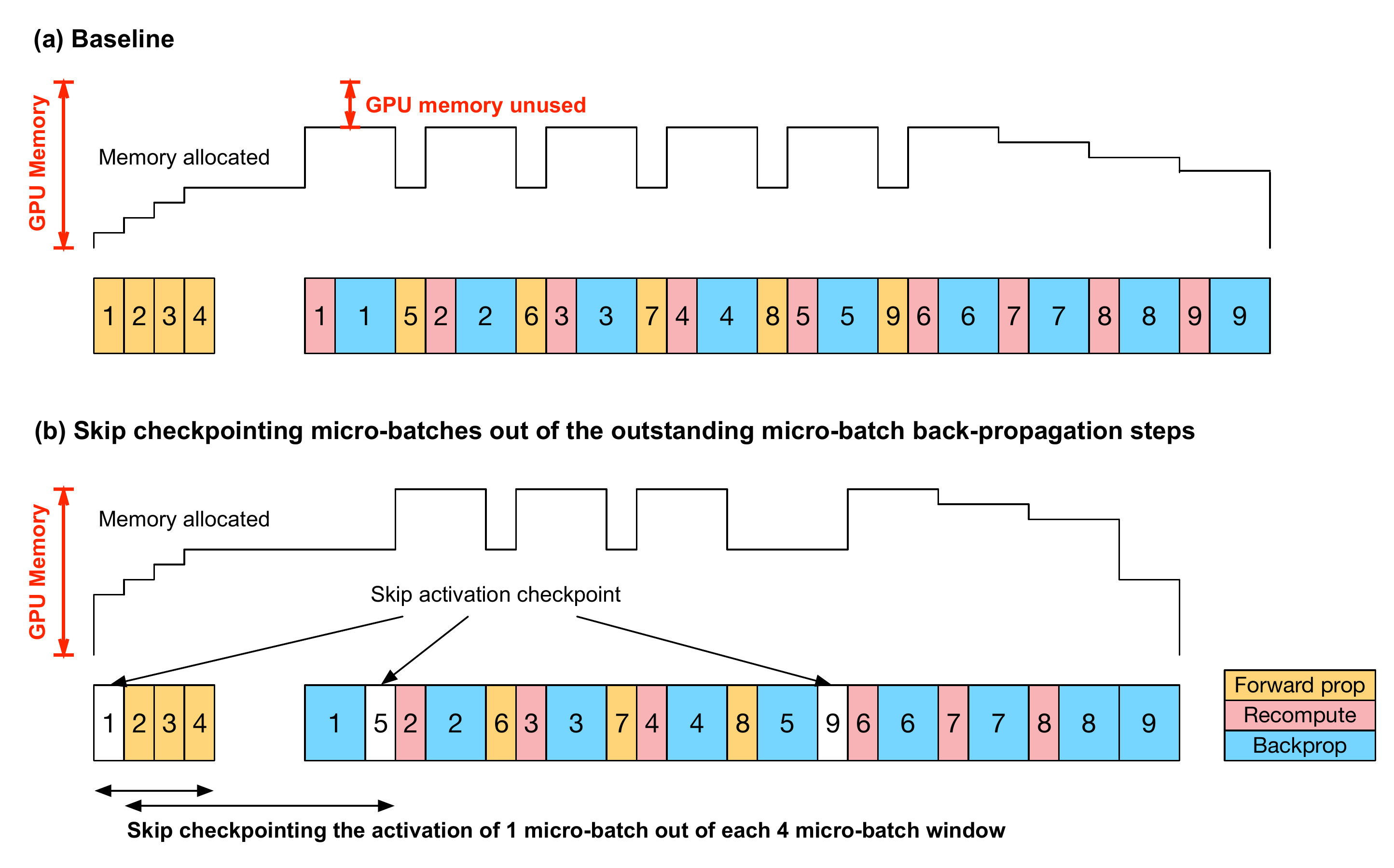}
  \caption{Computation and memory usage patterns of the baseline activation recomputation and microbatch level activation recomputation. Yellow boxes are a forward pass with activations checkpointed (i.e. only some activations are saved), red boxes are activation recomputation, blue boxes are backpropagation, and white boxes are a foward pass with all activations saved.}
  \label{fig:partial-activation-checkpoint}
\end{center}
\end{figure}

The computation and memory usage patterns of the first pipeline stage are illustrated in Figure~\ref{fig:partial-activation-checkpoint}. Yellow, red, and blue boxes indicate the execution of forward, recomputation, and back-propagation of one microbatch, respectively. This example uses the pipeline-parallel size of four and nine microbatches per iteration for training. Here the recomputation can be either full or selective. Figure~\ref{fig:partial-activation-checkpoint}.a depicts the execution flow of the baseline approach. Although there is some GPU memory left unused, if it is not large enough to store all of the activations and thus not require any recomputation, the activations of every microbatch is checkpointed and recomputed during back-propagation.

Microbatch level activation recomputation uses all the available device memory to store activations of some outstanding microbatches, and checkpoint the rest until the backpropagation frees enough memory to store another full layer of activations. We skip checkpointing microbatches (i.e. store all their activation) until all device memory is used. In Figure~\ref{fig:partial-activation-checkpoint}.b, the activations of the first microbatch are stored out of four microbatches. Once the activations of the first microbatch are consumed by the its back-propagation, enough memory is now free to store all activations of the fifth microbatch. This can be seen as a moving window of microbatches in which a certain number of them can have all activations stored.

Later pipeline stages have fewer outstanding backpropagation steps thus we can save all activations for the same number of microbatches within a smaller microbatch window. The size of outstanding microbatch backpropagation steps at each pipeline stage is calculated as $\max(0, p - S)$, where $S$ is the stage number. Based on our observations, many of later pipeline stages do not need any activation recomputation.

Microbatch level activation recomputation increases the model FLOPs utilization of the 175B and 530B parameter models to 52.3\% (+0.7\%) and 56.4\% (+0.4\%), respectively, compared to the baseline with both sequence parallelism and selective activation recomputation. The gain is small because the selective recomputation overhead is as small as $\sim$2\%. However, one can imagine model parallel configurations where there is nearly enough memory for no recomputation, in which case microbatch level activation recomputation could provide more improvement to training speed.

\end{document}